\pdfoutput=1
\documentclass[11pt]{article}

\usepackage{acl}

\usepackage{booktabs}
\usepackage{graphicx}
\usepackage{fancyvrb}
\usepackage{tablefootnote}
\usepackage{multirow}
\usepackage{paralist}

\usepackage{times}
\usepackage{latexsym}

\usepackage[T1]{fontenc}
\usepackage[utf8]{inputenc}
\usepackage{csquotes}
\usepackage{microtype}

\usepackage{inconsolata}

\title{Detection of Non-recorded Word Senses in English and Swedish}

 \author{Jonathan Lautenschlager \\ University of Stuttgart \\ \texttt{st167905@stud.uni-stuttgart.de}
 \And
 Emma Sk\"{o}ldberg \\ University of Gothenburg \\ \texttt{emma.skoldberg@svenska.gu.se}
 \AND
 Simon Hengchen\\ iguanodon.ai \& University de Genève\\ \texttt{simon@iguanodon.ai}
\And
 Dominik Schlechtweg\\ IMS, University of Stuttgart \\ \texttt{schlecdk@ims.uni-stuttgart.de}
}
\begin{document}
\maketitle

\begin{abstract}
This study addresses the task of Unknown Sense Detection in English and Swedish. The primary objective of this task is to determine whether the meaning of a particular word usage is documented in a dictionary or not. For this purpose, sense entries are compared with word usages from modern and historical corpora using a pre-trained Word-in-Context embedder that allows us to model this task in a few-shot scenario. Additionally, we use human annotations on the target corpora to adapt hyperparameters and evaluate our models using 5-fold cross-validation. Compared to a random sample from a corpus, our model is able to considerably increase the detected number of word usages with non-recorded senses.
\end{abstract}

\section{Introduction}

Dictionaries cover the senses of words at a particular point of time.
As they store vast amounts of lexical information in a convenient way, a lot of Natural Language Processing (NLP) tasks heavily rely on their quality and completeness. 
When a word gains a new sense or loses an old one in a speaker community, its dictionary entry may become outdated.
Lexicographers regularly check dictionaries for such outdated entries, a \enquote{challenge in lexicography over and above identifying new words themselves} \citep{Lau12p591}.

In this paper, we investigate systems that discover missing dictionary sense entries in modern English and Swedish dictionaries by comparing target word usages from reference corpora to the dictionary entries for the target word.
The basic task to solve is to decide whether the sense of a given word usage is covered by any sense in the dictionary entry of the target word or not.
The major difficulty is that a dictionary entry, despite containing a good amount of lexical information, does not provide enough context in order to train a standard Word Sense Disambiguation (WSD) system. 
Depending on the dictionary, the senses of an entry are sometimes only covered by the definition or a few example sentences.
We are addressing this problem by using a pre-trained Word-in-Context (WiC) model to create contextualized word embeddings \citep{cassotti-etal-2023-xl} and further comparing different ways of maximizing the quality of information we utilize from the dictionary entries.
Although we are not the first to address this problem \citep{outlier_detection, Lau12p591}, none of the related approaches utilizes WiC models to solve the task in a realistic scenario: \citeauthor{outlier_detection} does not operate in a few-shot scenario while \citeauthor{Lau12p591} do not automatically compare induced senses to a dictionary (but only between corpora).
Moreover, our model can even be applied to multilingual applications. 
Furthermore, we use a human annotation approach for a high-quality evaluation of our model's performance.

\section{Tasks \& Related Work}\label{sec:RelatedWork}
Detecting non-recorded senses in word usages has, especially with a direct focus on the practical applicability of dictionary maintenance, not been studied extensively \citep{outlier_detection, Lau12p591}.
Also known as \textit{Unknown Sense Detection}, the task combines aspects of \textit{Word Sense Disambiguation} and \textit{Word Sense Discrimination}. 
In the following, we will introduce some of the underlying concepts required to model these tasks.

\subsection{Word Sense Disambiguation}\label{WSD}
Word Sense Disambiguation (WSD) is a classical, yet unsolved task in NLP and has been studied for decades \citep{Weaver49, resnik-yarowsky-1997-perspective}.
\citet{WSD_Survey} describes the task of WSD as ``computationally identifying the meaning of a word by its use in a particular context'' (p. 1).
A word sequence $s$ that contains a particular target word $w$, at the position $i$ we call a \textit{word usage} of $w$ and represent it by the tuple $u = (s,i)$.
\citet[p. 4]{WSD_Survey} defines a \textit{word sense} as a commonly accepted meaning of a word.
Which meaning is expressed in a particular word usage depends on the context of use.
Let $\Phi = \{\phi_1, \phi_2, \dots, \phi_k\}$ be a set of predefined senses and $f: U \rightarrow \Phi$ be a mapping from usages to senses given by a lexical resource.
The task is to find mapping $f$.
So by definition, WSD can be considered a multiclass classification task.

In order to correctly classify a word usage based on a predefined set of senses, the sense has to be part of the set in the first place.
Building such \textit{sense inventory} with good coverage is strongly dependent on the available data.
Traditional WSD models use large amounts of manually annotated data (multi-shot) to train one \textit{classifier} per word to disambiguate its word usages \citep{burchardt2009framenet}.
Since manually labeled data sets are very time- and resource-consuming to produce, they are usually of a limited size and do not cover the full lexicon of the language.
While these classifier models generally deliver good results on a restricted set of words, they quickly suffer from the problem of data sparseness \citep[p. 4, p. 16]{WSD_Survey}.
More realistic WSD models therefore try to reduce the dependency on manually labeled training data.
Building such models makes it possible to extract the little training data needed (few-shot) from limited lexical resources like a dictionary which maps headwords\footnote{In the remainder of this paper, we use `headword' when referring to an entry in a lexical resource, and `lemma' as the canonical form of a given word. The two concepts greatly overlap.} to sets of sense entries.
A sense entry is usually structured, providing a sense definition (gloss), a number of example usages of the headword in that sense and some meta information such as the etymology and part-of-speech (see Appendix~\ref{app:wordnet} for an example).
Early few-shot computational approaches to WSD make use of such dictionaries and try to disambiguate word usages by simply counting overlapping words between eligible sense definitions of the target word and sense definitions of nearby words \citep{wsd_cooccurence}.
Newer models often use pre-trained contextualized embedders \citep{devlin-etal-2019-bert, peters-etal-2018-deep} to calculate vector representations for target usages and senses (Section \ref{sec:ContEmbeddings}), which can then be compared, e.g. with a similarity metric \citep{SensEmBERT, Hu19, zero_shot_wsd}.
\citet{glossreader-lcsdiscovery} develop a model training only one classifier for all words in their data.
In this way, knowledge between words can be shared, e.g., from frequent to infrequent ones \citep{kågebäck2016word, chen-etal-2021-non}. 

\subsection{Word Sense Induction}
\label{WSI}
Word Sense Induction (WSI), also called Word Sense Discrimination, differs from WSD in that it does not assume a predefined sense inventory. 
It can be thought of as the fully unsupervised counterpart of WSD and aims to group word usages by sense, instead of classifying word usages \citep{schutze1998automatic, outlier_detection}.
WSI can be seen as a clustering task \citep[cf.][p. 48]{Diday1976}:
Let $U = \{u_1, u_2, \dots u_n\}$ be a set of word usages of a word $w$.
Let $\Psi = \{\psi_1, \psi_2 \dots \psi_k\}$, $k \leq |U|$ be a set of labels representing non-structured senses for $w$ and a mapping $f: U \rightarrow \Psi$ given by a lexical resource.
The elements of a resulting equivalence class $C_\psi$ can now be seen as different usages of a word $w$ with the same sense $\psi$.
$$C_\psi=\{u|f(u) = \psi\}$$
The task is to find the set of equivalence classes $C = \{C_\psi|\psi \in \Psi\}$.
WSI has applications in lexicography \citep{lau-EtAl:2014} and lexical semantic change detection \citep{Lau12p591, Laicher2021explaining, Martinc2020evolution}.

\subsection{Unknown Sense Detection}\label{sec:USD}
The task of Unknown Sense Detection (USD), as defined by \citet{outlier_detection}, combines aspects of WSD and WSI.
While there are predefined senses, the task is not to map word usages to these, but instead to find word usages whose meanings are not covered by the sense inventory.
It is of no further importance by which known sense (if any) a word usage is described \citep{outlier_detection}.
Therefore, it can be seen as a binary classification task:
Let $U = \{u_1, u_2, \dots u_n\}$ be a set of word usages of a word $w$.
Let $\Phi = \{\phi_1, \phi_2, \dots, \phi_k\}$ be a set of predefined senses and $f: U \rightarrow \Phi$ be a mapping from usages to senses given by a lexical resource.
Let further $g: U \rightarrow \{0, 1\}$ be a mapping such that $g(u) = 1$ iff $f$ is undefined and $g(u) = 0$ otherwise.
The task is to find mapping $g$. 
That is, word usages that are not covered by any entry in our sense inventory need to be assigned label $1$ while covered usages need to be assigned label $0$. Although WSD, WSI and USD share fundamental characteristics, there are significant differences between them, as summarized in Table \ref{tab:task_differences}. Related tasks are Novel Sense Detection \citep[][]{Lau12p591,Jana2020novel} and Lexical Semantic Change Detection \citep[][]{schlechtweg-etal-2020-semeval,Kurtyigit2021discovery}, with the difference to USD being that they do not assume the existence of a dictionary, but rely on the comparison of corpora.

One of the few modeling approaches to USD is given by \citet{outlier_detection}, measuring the distances between the vector representations of target usages and sense entries.
If the target usage deviates far from all sense entries, i.e., for all sense entries another sense entry is closer than the target usage, it is considered an outlier \citep{outlier_detection}.

We go beyond previous work on USD by (i) evaluating models in a large-scale, realistic and practical scenario, (ii) testing state-of-the-art contextual embedders for solving USD and (iii) testing various heuristics to improve embedders for USD. As a result, we are the first to convincingly show that automatic methods help to update dictionaries.

\begin{table}[t]
\footnotesize
    \centering
    \setlength{\tabcolsep}{3pt}
    \begin{tabular}{llll}
    \toprule
        & \textbf{Task}                       & \textbf{Senses} & \textbf{Supervised}  \\
        \midrule
        WSD & Multiclass classification           & predefined               & yes              \\
        WSI & Clustering                          & none                     & no               \\
        USD & Binary classification               & predefined               & yes              \\
        \bottomrule
    \end{tabular}
      \caption{Differences between WSD, WSI and USD}
    \label{tab:task_differences}
\vspace{-0.5cm}
\end{table}

\subsection{Contextualized Word Meaning Representations}\label{sec:ContEmbeddings}
So-called \textit{contextualized embeddings} are numeric, high-dimensional vector representations of word meaning.
These are usually learned as parameters in a \textit{language model} trained on large amounts of data \citep[eg ELMO and BERT:][]{peters2018deep,devlin-etal-2019-bert}.
They encode distributional information \citep{Harris1954, firth1957synopsis} for words in context and can be used to measure the semantic similarity between word usages \citep[cf.][]{schutze1998automatic}.

BERT is one such contextualized embedding model that generates a high-dimensional vector representation for a given context.
The model architecture is based on the original multi-layer bidirectional Transformer cross-encoder by \citet{vaswani2017attention}.
BERT-embeddings find use in various NLP tasks \citep{devlin-etal-2019-bert}.
SentenceBERT (SBERT) is a modification of BERT overcoming limitations of cross-encoders using a bi-encoder in a Siamese Network, i.e., input sentences are encoded separately while the
weights of the underlying networks are shared \citep{reimers2019sentencebert}.
The recent XL-LEXEME model \citep{cassotti-etal-2023-xl} adapts the SBERT architecture to the Word-in-Context task \citep[WiC,][]{pilehvar2019wic} by giving prominence to the target word enclosing it into special tokens. The model was trained to minimize the contrastive loss with cosine distance on the multilingual WiC dataset \citep{pilehvar2019wic}, using XLMR \citep{conneau2019unsupervised} as the underlying language model.
It has been shown to perform extremely well on lexical semantic tasks such as lexical semantic change detection \citep{schlechtweg-etal-2020-semeval}.

\begin{table*}[t]
\footnotesize
\centering
\begin{tabular}{@{}lcccc@{}}
\toprule
\textbf{}          & \multicolumn{2}{c}{\textbf{Modern}} & \multicolumn{2}{c}{\textbf{Historical}} \\ \midrule
\textbf{Language}  & \textbf{English} & \textbf{Swedish} & \textbf{English}    & \textbf{Swedish}  \\
\textbf{Name}      & Leipzig\_News    & Leipzig\_News    & CCOHA               & Kubhist2          \\
\textbf{Year}      & 2020             & 2022             & 1810--1860          & 1790--1830        \\
\textbf{Source} &
  \multicolumn{2}{c}{\citet{goldhahn-etal-2012-building}} &
  \citet{alatrash-etal-2020-ccoha} &
  \citet{Kubhist} \\
\textbf{Sentences} & 1 million        & 1 million        & $\sim$250 thousand  & 3.3 million       \\ \bottomrule
\end{tabular}%
%    }
    \caption{Overview of corpora}
    \label{tab:corpora_overview}
\vspace{-0.5cm}
\end{table*}

\section{Datasets}
\subsection{Corpora}
\urldef{\WSleipzig}\url{https://cls.corpora.uni-leipzig.de/}
Since our goal is to support dictionary maintenance, we have to make sure to create a realistic scenario.
Use of language changes over time, which necessarily leads to the emergence of sense entry gaps or outdated sense entries in dictionaries.
Unrecorded word senses can occur for two main reasons: either they are old senses that were left unrecorded when the dictionary was created, or they are novel senses that emerged due to language change.
The former are likely to be found in historical data while the novel senses more likely occur in modern corpora.
To evaluate our models on both cases, we decided to use modern and historical corpora. 
Table \ref{tab:corpora_overview} shows an overview of our corpora. A precise statistical description of the modern corpora can be found on the website of Wortschatz Leipzig.\footnote{\WSleipzig}

\subsubsection{Modern}
We use the latest \textit{News} datasets from the Leipzig Corpora Collection \citep{goldhahn-etal-2012-building} for both English and Swedish.
These datasets contain each 1 million sentences in randomized order retrieved from articles on news websites published in the years 2020 (English) and 2022 (Swedish), without further processing.

\subsubsection{Historical}
We use the respective first corpus of the test data for the SemEval 2020 Task 1 \citep{schlechtweg-etal-2020-semeval} for English and Swedish.
The English corpus is based on CCOHA \citep{alatrash-etal-2020-ccoha}, a cleaned version of the well-known diachronic COHA corpus of American English \citep{davies2002corpus}.
The SemEval sample we use contains sentences in randomized order from the period 1810--1860 and consists of roughly 6 million tokens.
The Swedish corpus is a sample of KubHist2 \citep{Kubhist} containing sentences in randomized order from the period 1790--1830. It consists of about 71 million tokens, but contains frequent OCR errors \citep{DBLP:confdhnAdesamDT19}.

\subsection{Lexical resources}

WordNet 3.0 \citep{miller-1994-wordnet, fellbaum2005wordnet} is a well-known and well-established large lexical database for the English language.
It groups words which are synonymous in one of their meanings into so-called \textit{synsets} that describe distinct concepts or senses.
Each synset has one \textit{primary headword} under which the synset can be listed in a dictionary-like structure.
We refer to these as \textit{primary headwords}.
The word \textit{car}, for example, has five senses assigned, i.e., is part of five synsets, including the following two:
\begin{itemize}
  \setlength\itemsep{0em}
  \item[] \texttt{\textless\textbf{car}, auto, automobile, machine, motorcar\textgreater}
        \item[] \texttt{\textless{}cable car, \textbf{car}\textgreater}
\end{itemize}
The first synset, defined by its gloss as ``a motor vehicle with four wheels; usually propelled by an internal combustion engine'', has \textit{car} as the primary headword, followed by four other headwords from WordNet.
It also includes the example usage \textit{he needs a car to get to work}.
The second synset, defined by its gloss as ``a conveyance for passengers or freight on a cable railway'' has \textit{cable car} as the primary headword while \textit{car} is only an additional headword.
It includes the example usage \textit{they took a cable car to the top of the mountain}.\\
There are a total of 117,000 synsets, all containing a gloss.
Only a portion also includes example usages that put a member of the synset into a usage context with the respective meaning.

There are two ways to restructure WordNet to transform it into a dictionary-like structure, i.e., subordinating senses to headwords:
(i) Assign synsets only to the primary headword and accept definition gaps in return, or (ii) assign synsets to all participating headwords, but end up with duplicate senses in the dictionary.
We have opted for a mixture of both, by taking all synsets into account, but keeping track of whether it is a synset that has the headword as primary headword or not.
If this is the case, we refer to them as \textit{primary synsets}.
In the following, we explicitly specify when only primary synsets are meant.

\urldef{\svenska}\url{https://svenska.se/so/}
For Swedish, we use a processed data dump of \textit{Svensk ordbok} (SO),\footnote{\svenska. Please refer to the Limitations section.} the main contemporary dictionary for the Swedish language, created and maintained by the University of Gothenburg \citep{sw_dict_SO}.
It is structured in the classic style of a dictionary in that it lists senses under a headword.
In order to use a standardized form for such a headword, \citet{sw_dict_SO} applies the \textit{lemma-lexeme model} and groups different inflections and forms of words by their lemma.
In this way, our processed data dump of SO stores a total of 68,000 senses for over 41,500 headwords.
The majority of the senses are described by a \textit{gloss}, in some cases extended by a \textit{secondary gloss}.\footnote{We only made use of the secondary gloss if no standard gloss was given. see Appendix \ref{app:gloss} for an example of a secondary gloss.}
All senses are followed by example usages illustrating how the headword is used.

The two dictionaries\footnote{We use the term ``dictionary'' for WordNet here and in the remainder of this paper after having transformed it into a dictionary-like structure (see above).} represent the senses of words, describe them using both gloss and example usages, and group them by their headword.
Both dictionaries differ in that their entries have gaps in different parts of their sense entries.
While WordNet senses are fully covered by gloss but only have partial example usages, it is the other way around in the Swedish dictionary. They further differ in the way they treat historical senses: While WordNet only records recent senses, the Swedish dictionary covers entire modern Swedish period from 1521 onwards. Hence, the latter can be expected to cover historical senses, which fell out of use, to a much larger degree.
A direct comparison of the statistics of both dictionaries is shown in Table \ref{table:dictionaries}.

\begin{table}[t]
    \centering
    \small
    \begin{tabular}{@{}lrr@{}}
        \toprule
           & \textbf{WordNet} & \textbf{SO} \\ \midrule
        \textbf{Headwords}                                     & 86,555  & 41,597             \\
        \textbf{Avg. senses per headword}                      & 1.36    & 1.64               \\
        \textbf{Avg. senses per headword w.m.s.} & 2.27    & 2.91               \\
        \textbf{Percentage of senses with gloss}          & 100\%   & 79\%               \\
        \textbf{Avg. length of gloss}                     & 56.40   & 34.28              \\
        \textbf{Percentage of senses with examples}            & 28\%    & 100\%              \\
        \textbf{Avg. number of examples per sense}             & 0.41    & 3.37               \\
        \textbf{Avg. examples per sense w.e.}         & 2.84    & 3.37               \\
        \textbf{Avg. length of examples (char.)}                       & 33.60   & 32.44              \\ \bottomrule
    \end{tabular}
\caption{Direct comparison of the two dictionaries. Avg. senses per headword w.m.s. = Avg. senses per headword with multiple senses; Avg. examples per sense w.e. = Avg. examples per sense with examples}
\label{table:dictionaries}
\vspace{-0.3cm}
\end{table}

\section{Models}\label{sec:Models}
All our models have the same basic structure inspired by \citet{SensEmBERT}, but differ in the type of data used and the embedding comparison methodology.
Essentially, they first use the information given in the dictionary to create a vector representation for each existing sense.
The type of information used (e.g., gloss, example usage) and how it is processed (e.g., inserting headwords if missing) are treated as model hyperparameters, as listed below.
The models then create a usage embedding for the word usage to be examined and calculate the similarity to all sense embeddings of the associated headword.
If the calculated similarity score between the usage embedding and the closest sense embedding is below a specific threshold, the model predicts that the word usage is the occurrence of an unknown sense and therefore marks it as unassigned.\footnote{Find our code at: \url{https://github.com/ChangeIsKey/non-recorded-sense-detection}.} 
We decided to use XL-LEXEME (see Section \ref{sec:RelatedWork}) to vectorize both target usages and senses because of its good performance in Lexical Semantic Change Detection. This task involves the detection of word sense changes and thus XL-LEXEME can be expected to encode word sense information.

Each model's identifier is made up of a different choice of the hyperparameters, which will be explained in more detail in the following paragraphs:\footnote{The total computational cost of this hyperparameter search is roughly 100 hours of GPU time.}
\begin{itemize} 
  \setlength\itemsep{0em}
    \item usage embedding $\in$ \texttt{[$\epsilon$, SUB]},
    \item sense embedding $\in$ \texttt{[G0, G1, G2, G3, E0, E1, E2, E3, E4]},
    \item similarity measure $\in$ \texttt{[COS, SPR]},
    \item threshold $\in$ \texttt{$\{0.0, 0.01, 0.02, \dots, 1.0\}$}
\end{itemize}

\paragraph{Target Usage Embedding.}
We use XL-LEXEME to generate a contextualized embedding from each target usage and the position of the target word (hyper-parameter $\epsilon$).
The usages are obtained by directly searching for variations of the headwords (Section \ref{sec:annotation}), thus a headword must be contained in the usage in some inflected form. Some contextualized embedders are influenced by orthographic differences in target word form \citep{Laicher2021explaining}. Hence, we also test a variation in which the target word is replaced by the headword (\texttt{SUB}).

\paragraph{Sense Embedding.}
As for the target usages, we use the XL-LEXEME model to create contextualized embeddings representing the senses in our sense inventory.
The model encloses the target word in the word usage with special symbols.
However, glosses in dictionaries typically do not include the actual headword itself and therefore do not provide a target position.
A similar problem can occur with example usages:
In WordNet, the example usages of a synset always contain one of the headwords from its synset, but not necessarily the primary headword.
Therefore, not even the example usages are suitable without restriction.
Using various strategies, we thus modify the glosses and example usages if necessary. 
In Table \ref{tab:replacement_strategies}, all strategies are explained using an example usage taken from WordNet for a \textit{primary synset} of the headword \textit{inadequate}.
Instead of the headword itself, the example usage contains the headword \textit{poor}, which is also part of the synset. By applying these replacement strategies on both example usages and gloss, we receive the gloss models \texttt{G[0-3]} and the example models \texttt{E[0-4]} and \texttt{E[0-3]} for English and Swedish respectively.
The former represent a sense only by its gloss while the latter represent it only by its example usages.
In the case of multiple example usages we take their average embedding. 
Note that \texttt{E4} is only used for English, as Swedish example usages always contain the primary headword.

The models are naturally limited in their predictions by gaps in the data, i.e., missing gloss or example usages.
If a headword has no glosses or no example usages for all its senses, it is not represented by the sense inventory of the corresponding model and can therefore not be examined. 
Headwords where only individual senses are not covered by the data are part of the sense inventory.
However, the predictions are only of limited practical relevance since word usages of senses that are actually represented but have incomplete entries are predicted as unassigned as one cannot compare a word usage to a sense entry that has no information.

\newcounter{mpFootnoteValueSaver}
\begin{table}[t]
\small
\centering
\begin{tabular}{@{}lll@{}}
\toprule
 & \textbf{pattern} & \textbf{example} \\ \midrule
\textbf{0} & as is & a poor salary \\
\textbf{1} & HW: SQ & inadequate: a poor salary \\
\textbf{2} & SQ (HW) & a poor salary (inadequate) \\
\textbf{3} & SQ, i.e., HW & a poor salary, i.e., inadequate \\
\textbf{4} & replace word & an inadequate salary \\ \bottomrule
\end{tabular}%
\caption{Replacement strategies on an example of \textit{inadequate} taken from WordNet. ``dvs.'' is used in Swedish instead of ``i.e.'' for strategy 3, and strategy 2 in place of strategy 4 if there is no headword in the synset present. HW = HEADWORD; SQ = SEQUENCE}
\label{tab:replacement_strategies}
\vspace{-0.5cm}
\end{table}

\paragraph{Embedding Comparison.}
The similarity between the target embedding and each sense embedding is calculated using either Cosine Similarity \citep{salton1986introduction} or Spearman's rank correlation coefficient \citep{Spearman1904Association}. The latter shows better performance than Cosine Similarity for the WiC task in \citep{tabasi-etal-2022-exploiting}.
A threshold decides between \textit{assigned} and \textit{unassigned} word usages based on the similarity scores.
We test values in the range of $\{0.0, 0.01, 0.02, \dots, 1.0\}$.

\section{Annotation}
\label{sec:annotation}
We carry out two phases of human annotation.
The first phase is conducted both in order to collect data for the tuning of our models and to establish a reference point for the second annotation (random baseline).
The second phase serves the evaluation of the quality of our models' predictions.\footnote{Find the annotated data at: \url{https://zenodo.org/records/10718859}.}
 
The annotators are presented instances consisting of a word usage where the target word is marked and a sense gloss that may or may not describe this usage.
Additionally, they are shown the list of all possible sense glosses for the target word.
The annotation approach takes inspiration from \citet{Erk13}'s WSsim method by conducting an individual assessment of all senses for a usage, but differs in that we ask annotators for a binary classification into \textit{sense gloss fits} (label ``1'') and \textit{sense gloss does not fit} (label ``0''), instead of a five-level rating.
The annotators also always have the option to choose that no specification is possible (label ``-'').
They are encouraged to leave a comment if this is the case. An example of an annotation instance for a usage of the word \textit{relative} can be found in Appendix \ref{app:example_annotation_instance}.

\urldef{\trainingsdata}\url{https://github.com/ChangeIsKey/annotation_standardization/tree/main/use_single/wsbest/english/tutorial}
\urldef{\phitag}\url{https://phitag.ims.uni-stuttgart.de/}
\urldef{\Github}\url{https://github.com/joni0700/non-recorded-sense-detection}

A total of six annotators, three for each language, are recruited.
All annotators are students\footnote{At the exception of one annotator for Swedish.} and native speakers of the respective language. They are paid a fair wage in accordance with the laws of Sweden and Germany, respectively.
Before the annotation, they receive a 30-minute briefing during which they also conduct a short test annotation to familiarize with the process.
Both rounds of annotation are carried out using the PhiTag platform.\footnote{\phitag}

\subsection{Phase I: Random sample}\label{sec:annotationrandom}
In this phase, the annotation is conducted on a random sample of usages from the corpora.
Data sampling and briefing the annotators is carried out analogously for both languages.
Word usages from the modern and historical corpora are retrieved by lemmatizing the sentences and searching for appearances of a randomly selected subset of headwords from the language's dictionary.

In each corpus, we search word usages of a random sample of 3,000 headwords until we find at least one usage for 150 of them.
We keep at most five usages for each headword, chosen at random.
Lastly, we combine the word usages from the modern and the historical samples.
This results in a set of word usages that is approximately equally distributed between modern and historical corpora, which can be linked to a headword in the dictionary.
Combining each word usage with each eligible sense returns approximately 1,200 annotation instances, distributed over 500--700 usages, as listed in Table \ref{tab:HA_stats1}.
These are assessed by the three annotators.
For WordNet, only primary synsets are considered in the first annotation phase (see also Section \ref{sec:limitations}).

Find a summary of the annotation results in Table \ref{tab:HA_stats1}. We aggregate judgments per annotation instance by majority.
Then, we aggregate these majority labels by usage to decide whether the usage is assigned or unassigned:
A usage is considered assigned iff at least one instance has the majority label ``1''. 
For English, out of 473 usages, 45 are labeled unassigned (9.5\%), i.e., all their instances have the majority label ``0''.
For Swedish, out of 674 usages, 95 are labeled unassigned (16.6\%).

We report inter-annotator agreement scores measured by Krippendorf's alpha \citep{krippendorff2018content} in Table \ref{tab:HA1_IAA} in Appendix \ref{app:iaa}. Agreement is low to moderate: In English it ranges from 0.273 to 0.561, while Swedish has more agreement (minimum 0.317, maximum 0.655).
The pairwise comparison reveals that annotator A1 of English stands out in particular and the observation holds for both modern and historical usages.\footnote{This annotator was excluded from the second annotation phase.}
For Swedish, the annotators are more in agreement, especially on the modern data.

\subsection{Phase II: Model prediction}\label{sec:annotationpredictions}
We predict word usages with non-recorded senses as described below in Section \ref{sec:expprediction}. These usages are then uploaded to PhiTag and annotated in parallel to the first phase. One English annotator from the first phase is excluded for judgment inconsistencies and a new annotator is recruited.

Find a summary of the annotation results in Table \ref{tab:HA_stats2}. In comparison to the first phase (see column ``Unassigned baseline''), the second phase yields considerably higher shares of unassigned usages for both languages, which will be discussed in more detail in Section \ref{sec:expprediction}.

For English, the annotator-agreement on the full data is 0.384, as displayed in Table \ref{tab:HA2_IAA} in Appendix \ref{app:iaa}.
The agreement is comparable to the first phase. However, note that we increased the number of senses the annotators can choose from by including the number of non-primary synsets.
There is considerably higher consensus among the Swedish annotators (0.56).
For English, agreement on historical instances is slightly higher than on the modern instances.
This contrasts with phase I.
For Swedish, it is considerably worse on historical instances, as in phase I.

\section{Experiments}

\subsection{Phase I: Model selection}\label{sec:expselection}

We select hyper-parameters for our models on the randomly sampled and annotated data from Phase I (see Section \ref{sec:annotationrandom}). In order not to overfit them on the limited data, we assess them using k-fold cross-validation.
We perform a total of 10 rounds of 5-fold cross-validation for every model.
Each round consists of different manipulations of the gold sense assignments.
In each round, we randomly mask a set of assigned senses as unassigned in order to simulate unknown senses \citep[cf. the method described in][]{outlier_detection} because the number the number of naturally unassigned senses was very low.\footnote{Unassigned senses are left untouched and are included in the evaluation data.}
Before masking, we remove incomplete (i.e., not providing sufficient information) senses entries. Senses that are the only complete sense of a headword are always unmasked as masking them would exclude the whole target word from evaluation.
We then randomly select one sense for each headword to be unmasked and mask the rest.\footnote{We use the same simulated data for all models of a language, even though the gloss models in English and the example models in Swedish do not suffer from limited dictionary completeness.} 
Find an example of this procedure described in Appendix \ref{app:example_prediction_instance}.

Based on this random masking, we label all word usages from the human annotation \textit{assigned} (label ``0'') if at least one assigned sense is not masked.
All other word usages are labeled \textit{unassigned} (label ``1'').
This set of word usages is then randomly divided into 5 folds for the cross-validation.  

Applying the randomization ten times leaves us with ten different data sets of 5-fold subdivided usages.
In each round for each fold $k$, the model uses the data of the remaining four \textit{training}-folds to determine a threshold $\in \{0.0, 0.01, 0.02, \dots, 1.0\}$ (see Section \ref{sec:Models}) for the respective similarity measure, so that the $F_{0.3}$-score is maximized, giving more importance to precision over recall.\footnote{Find a discussion of the evaluation metric selection in Appendix \ref{sec:metric}.}
Using this threshold, the model then predicts on the \textit{test}-fold $k$. Find a detailed example of our evaluation pipeline in Appendix \ref{app:example_prediction_instance}.

In addition, we establish baselines to be able to better understand the performance.
This includes a random baseline that predicts \textit{assigned} with a probability of $p$ where $p$ is equal to the share of assigned usages in the gold data,\footnote{This is an approximation of a majority baseline. Given our majority class being label 0 (assigned), a normal majority baseline is not applicable as the F-score would be undefined.} and a frequency-baseline that only predicts \textit{assigned} if a usage can be assigned to its most frequent sense and unassigned otherwise. 
This baseline only applies to WordNet where we have sense frequency information.

\paragraph{Results.} When optimizing for $F_{0.3}$-scores, model \texttt{E4\_COS} produces the best results for English in the cross validation (see Appendix~\ref{app:cv-3-results} for an overview and Appendix~\ref{app:cvexample} for an example of an individual evaluation round).
However, since the value deviates only minimally from the performance of \texttt{E4\_SUB\_SPR}, we decide to use the latter, as the standard deviation of the scores over ten rounds of cross-validation is lower.
For Swedish, we decide on \texttt{G3\_COS}, as it achieves the best results. All average model performances exceed the baselines.

\begin{table}[t]
\footnotesize
\centering
    \begin{tabular}{@{}lrr@{}}
    \toprule
     & \textbf{English} & \textbf{Swedish}  \\ \midrule
    \textbf{Instances} & 1165 & 1202       \\
    \textbf{Usages} & 474 & 706   \\
    \textbf{Label dist. (0, 1, -)} & (1840, 1651, 4) & (1294, 2104, 208) \\
    \textbf{Excluded instances} & 2 & 87  \\
    \textbf{Remaining usages} & 473 & 674 \\
    \textbf{Assigned} & 428 & 562 \\
    \textbf{Unassigned} & 45 (9.5\%) & 95 (16.6\%) \\ \bottomrule
    \end{tabular}
\caption{Statistics of annotation phase I (random sample)}
\label{tab:HA_stats1}
\vspace{-0.5cm}
\end{table}

\begin{table}[t]
\footnotesize
\centering
    \begin{tabular}{@{}lrr@{}}
    \toprule
     & \textbf{English} & \textbf{Swedish}  \\ \midrule
    \textbf{Instances}  & 1208             & 1400              \\
    \textbf{Usages}  & 322              & 1001              \\
    \textbf{Label dist. (0, 1, -)}  & (2151, 1462, 11) & (2529, 1218, 456) \\
    \textbf{Excluded instances} & 5                & 109               \\
    \textbf{Remaining usages} & 322              & 927               \\
    \textbf{Assigned} & 277              & 327               \\
    \textbf{Unassigned} & 45 (13.98\%)     & 600 (64.725\%)    \\ \hline
    \textbf{Unassigned baseline} & 45 (9.5\%) & 95 (16.6\%) \\ \bottomrule
    \end{tabular}
\caption{Statistics of annotation phase II (model prediction). The bottom row gives the annotation result from phase I obtained from randomly sampling usages as a baseline.}
\label{tab:HA_stats2}
\vspace{-0.5cm}
\end{table}

\subsection{Phase II: Model prediction}\label{sec:expprediction}

The two models selected above predict on equally-sized subsets of 100--150K sentences, for both modern and historical corpora, respectively.
The sentences are cleaned and then filtered by excluding those that are longer than 300 characters or have too many punctuation tokens ($> 25\%$).
Every sentence is lemmatized and then searched for the lemmas of the headwords represented in the respective model's sense inventory.
For English, this time not only primary synsets are considered in this sense inventory (in contrast to phase I) in order to prevent missing senses recorded under another headword.
(We consider this difference to the first round in the evaluation below.)
When a headword is found, the model predicts for this usage, whether it is assigned or not.
The resulting prediction sample consists of 3,608 usages and 706 headwords in English and 12,534 usages and 2,061 headwords in Swedish.
In a second step, we further process this sample in the following ways:
\begin{itemize}
  \setlength\itemsep{0em}
    \item We exclude partially complete headwords (at least one sense incomplete).
    \item Unassigned usages are sorted by similarity to the nearest sense, i.e., the usage that is the least similar to all eligible senses is first.
\end{itemize}
Then, usages for evaluation are chosen from the top of this list.
At most eight usages for the same headword are sampled to ensure a broad analysis.
These usages are combined with all eligible senses from the respective dictionary.
The finally resulting sample has roughly the same size as in the first annotation phase. 
This is annotated in parallel to the first phase. The annotation analysis is described above in Section \ref{sec:annotationpredictions}.

\paragraph{Results.}
Table \ref{tab:HA_stats2} summarizes the annotation results from phase II, yielding diverse results.
While both models successfully increase the number of unassigned usages in the sample data compared to the first phase (random sample, see column ``Unassigned baseline''), the Swedish model is considerably more successful: Out of 1000 predictions, the Swedish model is correct in almost two thirds of cases according to human judgment.
In the predictions of the Swedish model, there is a significantly higher number of usages with few senses (1,400 instances for 1,001 usages vs. 1,208 instances for 322 usages, even though we have similar numbers of senses per headword for both dictionaries, as shown in Table \ref{table:dictionaries}).

Note that in English, the samples from both phases are not completely comparable because secondary synsets were excluded in the first phase.
However, taking them into account for the first phase would likely reduce the number of unassigned usages, which would only strengthen our case.

Differentiating between modern and historical instances indicates a commonality between both languages (see Appendix \ref{app:anncorpus}):
In the first phase, unassigned senses were almost equally distributed between modern and historical usages in the English data.
In the second phase, however, the proportion of correctly predicted historical usages hardly differs from the random sample of the first phase while the correct predictions of modern usages increases by two thirds.
We observe a similar trend in Swedish: Although the distribution of unassigned usages is very different between modern and historical in the first Swedish phase with proportionately almost twice as many unassigned usages in the historical data, a stronger relative increase in the modern usages can also be observed here.

\paragraph{Manual Analysis.}
We perform a manual analysis of the true positives from the model predictions.
For English, we found several cases where also a close manual analysis suggests that they are truly non-recorded in our dictionary.
As an example, consider the following metaphoric word usage of \textit{pipeline}:
\begin{itemize}
   \setlength\itemsep{0em}
   \item[] \textbf{usage}: There should be some things that can be done in the short term, but in terms of developing the \textbf{\textcolor{SpringGreen}{pipeline}} further on coaching and executive positions, that would take a longer period of time.
    \item[] \textbf{senses}: ``a pipe used to transport liquids or gases''; ``gossip spread by spoken communication''
\end{itemize}
Further examples include usages of \textit{qualification}, \textit{baked} or \textit{to blacken}. However, true positives also contain a number of problematic cases: a major problem are multi-word expressions.
Our word usages sampling algorithm has only very limited capability to detect multi-word expressions like e.g. \textit{revolves around} as it only detects the headword \textit{revolves}. Thus the model will compare the usage to the wrong sense entries, i.e., not the ones of the headword \textit{revolve around}.
Note, however, that there are also cases where multi-word expressions, like idioms, are involved, but no sense entry is present:
\textit{to carry the \textbf{flame} of reforms}, or \textit{to go at \textbf{length}}. Such usages classify as non-recorded senses.

\section{Conclusion}

Our goal was to automatically detect non-recorded word senses in historical and modern corpora based on a realistic sense inventory providing limited information. 
Our model uses a pre-trained Word-in-Context embedder to generate target usage and sense embeddings and decides, based on a threshold for similarity, whether any sense from the inventory is expressed by the usage. 
Our method considerably increases the chance to find non-recorded word senses in corpus usages compared to a random baseline (Phase II vs. Phase I).
We predict a large number of unassigned usages that can be used to update WordNet's and SO's sense inventory.
We observe that the models show different behavior on modern and historical data and identify a number of problems in the modeling pipeline that should be approached by future work. Most importantly, the detection of headword usages needs to be improved.
We believe that it is possible that our approach can be used in the near future to provide practical support in dictionary maintenance.

\section{Limitations}
\label{sec:limitations}
We only realised throughout the experiments for phase I that we had missed a number of senses that were recorded as secondary synsets. Hence, we decided to include these for phase II. Due to the exclusion of non-primary synsets in the first annotation phase, an unrestricted comparison between the two English annotation phases is not possible. To overcome this limitation, future experiments ought to ensure that pre- and post-prediction annotation conditions closely mirror each other. 

Moreover, and the baselines described in Section \ref{sec:expselection} were run independently for each model. Future work should fix these across models in order to make model performances more comparable.
Future work should also investigate the use of ensemble methods, or combinations that use both gloss and example representations (e.g., by averaging vectors).

Manual analysis showed that the model architecture has some weaknesses: faulty detection of headwords' word usages is the major problem. Additionally, a faulty regex in the data processing pipeline resulted in diacritics being stripped from usages for the prediction phase: As such, our models were at a disadvantage as they had to predict senses using dirty data (e.g. `säger' ``says'', from the verb ``to say'', became `sger'). This presumably impacted Swedish predictions much more than English ones.

Another limitation of our work is the treatment of SO: In our extraction from the SO database, we did not consider sub-senses of headwords (e.g. figurative uses) and only kept a word's \textit{main} senses. This led to an over-classification for non-recorded senses.\footnote{The code is available at \url{https://github.com/ChangeIsKey/SO-extract-db}. A handful of sense glosses were extracted with errors.}

\section*{Acknowledgments}
This paper is a revised version of \citet{Lautenschlager2023thesis} on which we carried out several additional analyses.
Dominik Schlechtweg and Emma Sk\"{o}ldberg have been funded by the research program `Change is Key!' supported by Riksbankens Jubileumsfond (under reference number M21-0021) during the creation of this study. We thank Haim Dubossarsky and Nina Tahmasebi for contributions to this project.
This work has been made possible by code created by iguanodon.ai. 

\bibliography{references,Bibliography-general,bibliography-self,bibliography-supervision-self}

\clearpage
\appendix

\onecolumn
\section{Wordnet in a dictionary form}
\label{app:wordnet}
This is the Wordnet data for the headword \textit{unable} in dictionary form.
\newsavebox\myv

\begin{lrbox}{\myv}\begin{minipage}{\textwidth}
\begin{verbatim}

    {
        "entries": [
            {
                "pos": "adj.",
                "gloss": "(usually followed by `to') not having the necessary means 
                    or skill or know-how",
                "examples": [
                    "unable to get to town without a car",
                    "unable to obtain funds"
                ]
            },
            {
                "pos": "s",
                "gloss": "(usually followed by `to') lacking necessary physical or 
                    mental ability",
                "examples": [
                    "dyslexics are unable to learn to read adequately",
                    "the sun was unable to melt enough snow"
                ]
            }
        ],
        "headword": "unable"
    },
\end{verbatim}
\end{minipage}\end{lrbox}

\resizebox{0.8\textwidth}{!}{\usebox\myv}\\
\clearpage
\section{Gloss and secondary gloss}
\label{app:gloss}
Find below the entry for the Swedish noun \emph{svindel}. The word has two main senses, listed in \texttt{definitions}. While the first sense is only described by a gloss, the second sense has both a main gloss (``(ekonomiskt) bedrägeri i större skala") and a secondary gloss (``i större skala").

\begin{lrbox}{\myv}\begin{minipage}{\textwidth}
\begin{verbatim}
     {
        "key": 145357,
        "word": "svindel",
        "nature": "subst.",
        "definitions": [
            {
                "gloss": "+yrsel som uppkommer vid vistelse på höga höjder",
                "sub_gloss": "",
                "sub_entries": [
                    {
                        "gloss": "äv. om likn. känsla som uppstått av annan orsak",
                        "sub_gloss": "",
                        "sub_entries": [],
                        "examples": [
                            "han kände svindel vid tanken på hur mycket pengar han hade ansvar för"
                        ],
                        "year": "1668"
                    }
                ],
                "examples": [
                    "hon fick svindel uppe i tornet"
                ],
                "year": "1668"
            },
            {
                "gloss": "(ekonomiskt) bedrägeri",
                "sub_gloss": "i större skala",
                "sub_entries": [],
                "examples": [],
                "year": "1879"
            }
        ]
    }
\end{verbatim}
\end{minipage}\end{lrbox}

\resizebox{0.8\textwidth}{!}{\usebox\myv}\\

\onecolumn
\section{Example of an annotation instance}
\label{app:example_annotation_instance}
Example of an annotation instance for a usage of the word \textit{relative}.
The word itself is highlighted, one gloss is marked for annotation and all glosses are given as additional information. The annotator is asked to choose between label ``1'' (\textit{sense gloss fits}) and label ``0'' (\textit{sense gloss does not fit}). The annotators also always have the option to choose that no specification is possible (label ``-'').
\begin{Verbatim}[commandchars=\\\{\}]
\textbf{usage:}
Abigail’s sister accompanied her to this meeting because the sister alleged that 
she had witnessed some of the childhood sexual abuse from the \textbf{\textcolor{SpringGreen}{relative}}.
    
\textbf{gloss:}
estimated by comparison; not absolute or complete
    
\textbf{all glosses:}
estimated by comparison; not absolute or complete
    
an animal or plant that bears a relationship to another
(as related by common descent or by membership in the same genus)
    
a person related by blood or marriage   
\end{Verbatim}

\section{Inter-annotator agreement}
\label{app:iaa}
\begin{table}[!h]
\footnotesize
\centering
\begin{tabular}{@{}lrrrrrr@{}}
\toprule
                   & \multicolumn{3}{c}{\textbf{English}}               & \multicolumn{3}{c}{\textbf{Swedish}}               \\ \midrule
                   & \textbf{All} & \textbf{Modern} & \textbf{Historic} & \textbf{All} & \textbf{Modern} & \textbf{Historic} \\
\textbf{A1 vs. A2} & 0.331        & 0.392           & 0.276             & 0.408        & 0.506           & 0.317             \\
\textbf{A1 vs. A3} & 0.273        & 0.333           & 0.220             & 0.575        & 0.614           & 0.527             \\
\textbf{A2 vs. A3} & 0.561        & 0.580           & 0.544             & 0.524        & 0.655           & 0.353             \\
\textbf{Full}      & 0.389        & 0.434           & 0.349             & 0.480        & 0.588           & 0.371             \\ \bottomrule
\end{tabular}
\caption{Inter-annotator agreement using Krippendorff's alpha of annotation phase I. The labels A1--A3 represent different annotators across languages. The label ``-'' is excluded from the data before calculating agreement}
\label{tab:HA1_IAA}
\end{table}

\begin{table*}[!h]
\footnotesize
\centering
\begin{tabular}{@{}lrrrrrr@{}}
\toprule
                   & \multicolumn{3}{c}{\textbf{English}}               & \multicolumn{3}{c}{\textbf{Swedish}}               \\ \midrule
                   & \textbf{All} & \textbf{Modern} & \textbf{Historic} & \textbf{All} & \textbf{Modern} & \textbf{Historic} \\
\textbf{A1 vs. A2} & 0.335        & 0.310           & 0.347             & 0.539        & 0.566           & 0.456             \\
\textbf{A1 vs. A3} & 0.342        & 0.291           & 0.397             & 0.558        & 0.546           & 0.552             \\
\textbf{A2 vs. A3} & 0.464        & 0.490           & 0.374             & 0.608        & 0.641           & 0.526             \\
\textbf{Full}      & 0.384        & 0.368           & 0.375             & 0.559        & 0.584           & 0.490             \\ \bottomrule
\end{tabular}
\caption{Inter-annotator agreement using Krippendorff's alpha of annotation phase II. The labels A1--A3 represent different annotators across languages. The label ``-'' is excluded from the data before calculating agreement}
\label{tab:HA2_IAA}
\end{table*}

\newpage 

\section{Example of evaluation pipeline}
\label{app:example_prediction_instance}
Consider the annotation instance for the word \textit{relative} from Appendix \ref{app:example_annotation_instance}. Annotators will likely assign this usage to the gloss ``a person related by blood or marriage''. Let's assume that the annotation outcome for this instance from the three annotators is \((1,1,-)\). These are aggregated by majority vote as described in Section \ref{sec:annotationrandom}, in this case $1$. Hence, this usage is considered assigned. For clarity, we further assume that annotations for all other glosses of this usage are aggregated as $0$, i.e., the usage is unambiguously assigned to the above-mentioned sense gloss. The annotation instance is then manipulated to perform model selection during the simulation (masking) and cross-validation procedure described in Section \ref{sec:expselection}. For each round of the simulation procedure, there are two possibilities: (i) the assigned sense is randomly masked, or (ii) not. For (i), models are optimized to predict the usage as unassigned (unknown), i.e., they should not assign this usage to any remaining (non-masked) glosses/senses of the word \textit{relative}. For (ii), models are optimized to predict the usage as assigned (known), i.e., the usage should be assigned to any of the glosses/senses. (Although we know the correct assignment, we do not distinguish between possible assignments.) Gloss models (G) create their prediction for each instance by taking the word usage and the list of possible glosses as input, manipulating these according to Table \ref{tab:replacement_strategies}, embedding usage and sense gloss, computing each pairwise cosine similarity and deciding on the assignment according to a threshold on the similarity. The threshold is optimized globally over all prediction instances. Example models (E) work similarly with the difference that they do not embed a sense according to their gloss, but according to observed examples of that sense.

\section{Selection of evaluation metric}\label{sec:metric}

The $F_\beta$-score modifies the more common $F$-score so that a recall factor $\beta$ changes the importance of the recall.
By choosing a $\beta < 1$, precision is rated more important.
The smaller the $\beta$ the more important becomes precision.
Precision is the share of true positives in the instances predicted by the model as positive.
Hence, optimizing for precision will increase the share of true positives (by decreasing the number of false positives) in our predictions.
Prioritizing precision will also tend to decrease the number of usages labeled by the model as unassigned because choosing the upper percentiles of a (meaningful) rank-cutoff-based classifier will tend to decrease the probability for a false positive, as opposed to including lower percentiles.
In terms of practical application, we deem a small yet precise sample as much more useful than a large, imprecise one.

\begin{figure}[h!]
    \centering
    \includegraphics[scale=0.9]{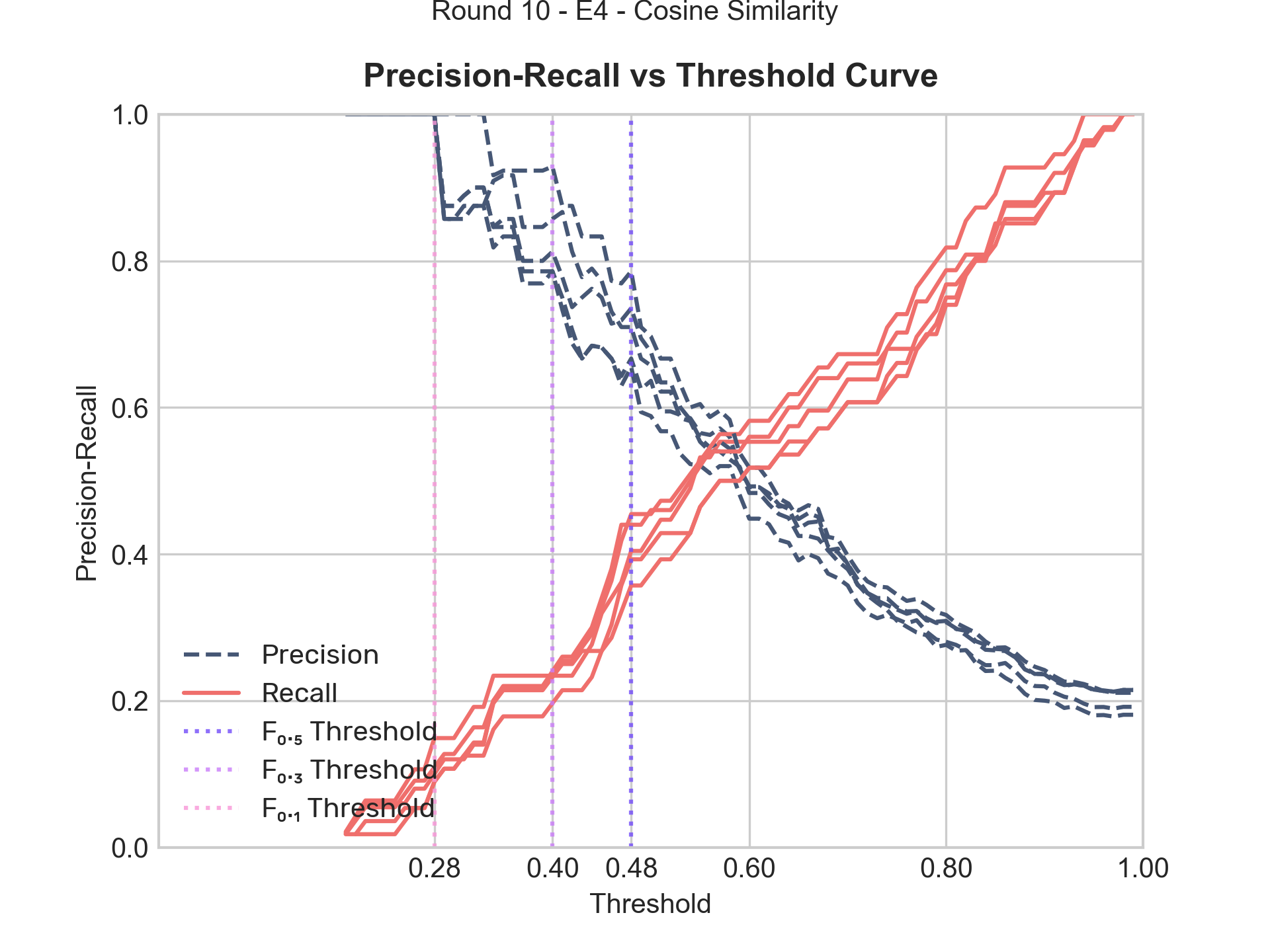}
    \caption{Precisions and recalls of all five folds in the cross-validation round 10 of model \texttt{E4\_COS} on English data}
    \label{fig:pr_t_curve}
\end{figure}

We perform the 5-fold cross-validation described in Section \ref{sec:expselection} with various $F_\beta$-scores maximizing for $\beta \in \{0.1, 0.3, 0.5\}$ separately to check model predictions with varying importance of precision. Find an example for the results of one round of cross-validation with $F_{0.3}$ in Table \ref{tab:cv_e4_r10}.
We calculate the average of all values for the test fold and the training folds for each round, respectively.
Figure \ref{fig:pr_t_curve} shows the precision-recall vs. threshold curve and nicely illustrates the differences between the different scores:
While $F_{0.1}$ accepts almost any recall in order to maximize the precision, $F_{0.3}$ and $F_{0.5}$ find a balance between the two values. 
After manually checking model predictions and analyzing performances we choose the $F_{0.3}$-score for model selection.
While $F_{0.1}$-score would be even better in terms of precision, we observe in the individual rounds of the cross-validation that a $\beta$ of $0.1$ is too extreme and excludes a disproportionate number of true positives.

\newpage

\section{Cross-validation overview}\label{app:cv-3-results}
\begin{table*}[!h]
\centering
\small
\begin{tabular}{@{}lrrrrrrrr@{}}
\toprule
\multicolumn{9}{c}{$F_{0.3}$-score} \\ \midrule
\multirow{3}{*}{\textbf{}} & \multicolumn{4}{c}{\textbf{English}} & \multicolumn{4}{c}{\textbf{Swedish}} \\
 & \multicolumn{2}{c}{\textbf{DEFAULT}} & \multicolumn{2}{c}{\textbf{SUB.}} & \multicolumn{2}{c}{\textbf{DEFAULT}} & \multicolumn{2}{c}{\textbf{SUB.}} \\
 & \textbf{COS} & \textbf{SPR} & \textbf{COS} & \textbf{SPR} & \textbf{COS} & \textbf{SPR} & \textbf{COS} & \textbf{SPR} \\
 \midrule
\textbf{E0} & 0.466 & 0.523 & 0.477 & 0.452 & 0.417 & 0.419 & 0.413 & 0.400 \\
\textbf{E1} & 0.589 & 0.590 & 0.583 & 0.592 & 0.451 & 0.443 & 0.413 & 0.399 \\
\textbf{E2} & 0.579 & 0.562 & 0.566 & 0.590 & 0.425 & 0.431 & 0.411 & 0.408 \\
\textbf{E3} & 0.494 & 0.523 & 0.493 & 0.489 & 0.428 & 0.431 & 0.397 & 0.392 \\
\textbf{E4} & \textbf{0.613} & 0.584 & 0.593 & \textbf{0.612} &  &  &  &  \\
\textbf{G0} & 0.270 & 0.280 & 0.255 & 0.267 & 0.349 & 0.371 & 0.345 & 0.340 \\
\textbf{G1} & 0.227 & 0.220 & 0.226 & 0.209 & 0.600 & 0.606 & 0.549 & 0.537 \\
\textbf{G2} & 0.260 & 0.223 & 0.264 & 0.245 & 0.550 & 0.612 & 0.564 & 0.547 \\
\textbf{G3} & 0.217 & 0.259 & 0.275 & 0.256 & \textbf{0.625} & 0.617 & 0.599 & 0.621 \\ \bottomrule
\end{tabular}
\caption{Results of cross-validation. Performance is given as average $F_{0.3}$ across rounds and folds}
\end{table*}
\clearpage
\section{Cross-validation in round 10 of model \texttt{E4\_COS} for English}
\label{app:cvexample}

\begin{table*}[!h]
\scriptsize
\centering
\begin{tabular}{lrrrrrrrrrrrr}

\hline
 &
  \multicolumn{2}{c}{\textbf{Average}}
  \textbf{} &
  \multicolumn{2}{c}{\textbf{Fold 1}}
   &
  \multicolumn{2}{c}{\textbf{Fold 2}}
  \textbf{} &
  \multicolumn{2}{c}{\textbf{Fold 3}}
  \textbf{} &
  \multicolumn{2}{c}{\textbf{Fold 4}}
  \textbf{} &
  \multicolumn{2}{c}{\textbf{Fold 5}}
  \textbf{} \\
 &
  \textbf{Training} &
  \textbf{Test} &
  \textbf{Training} &
  \textbf{Test} &
  \textbf{Training} &
  \textbf{Test} &
  \textbf{Training} &
  \textbf{Test} &
  \textbf{Training} &
  \textbf{Test} &
  \textbf{Training} &
  \textbf{Test} \\ \hline
\textbf{Threshold}        & 0.396      &      & 0.340      &      & 0.350      &      & 0.480      &      & 0.410      &      & 0.400      &   \\   
\textbf{Precision}        & 0.842      & 0.720      & 0.846      & 1.000      & 0.833      & 1.000      & 0.735      & 0.500      & 0.867      & 0.600      & 0.929      & 0.500      \\
\textbf{Recall}         & 0.272      & 0.215      & 0.234      & 0.105      & 0.179      & 0.400      & 0.455      & 0.182      & 0.260      & 0.188      & 0.232      & 0.200      \\
\textbf{$\mathbf{F_{0.3}}$}      & 0.701      & \textbf{0.573}      & 0.696      & 0.588      & 0.640      & \textbf{0.890}      & 0.700      & \textbf{0.437}      & 0.727      & 0.508      & 0.744      & \textbf{0.445}      \\
\textbf{random\_$\mathbf{F_{0.3}}$} & 0.335      & 0.331      & 0.328      & 0.336      & 0.328      & 0.400      & 0.358      & 0.000      & 0.328      & \textbf{0.528}      & 0.334      & 0.392      \\
\textbf{frequency\_$\mathbf{F_{0.3}}$} & 0.598      & 0.549      & 0.563      & \textbf{0.707}      & 0.544      & 0.790      & 0.607      & 0.365      & 0.649      & 0.440      & 0.628      & 0.445      \\ \hline
\end{tabular}
\caption{Cross-validation results in round 10 of model \texttt{E4\_COS} when maximizing $F_{0.3}$-score on English data. Column ``SUB.'' gives results for models where the target word was substituted with the headword}
\label{tab:cv_e4_r10}
\end{table*}

\section{Comparison of annotation results per corpus}
\label{app:anncorpus}
\begin{table*}[!h]
\centering
\footnotesize
\begin{tabular}{@{}lrrrrrr@{}}
\toprule
                    & \multicolumn{3}{c}{\textbf{~~~~~~~~~Phase I: Random sample}}                 & \multicolumn{3}{c}{\textbf{~~~~~~~~~~~~~~~~~~Phase II: Model prediction}}        \\
\textbf{Usages}     & \textbf{All} & \textbf{Modern} & \textbf{Historical} & \textbf{All} & \textbf{Modern} & \textbf{Historical} \\ \midrule
\textbf{total}     & 474 & 232 & 242 & 322 & 210 & 112 \\
\textbf{excluded}  & 1   & 1   & 0   & 0   & 0   & 0   \\
\textbf{remaining} & 473 & 231 & 242 & 322 & 210 & 112 \\
\textbf{assigned}  & 428 & 208 & 220 & 277 & 176 & 101 \\
\textbf{unassigned} & 45 (9.5\%)   & 23 (9.9\%)      & 22 (9.1\%)          & 45 (13.98\%) & 34 (16.2\%)     & 11 (9.8\%)    \\ \bottomrule     
\end{tabular}
\caption{Annotation results on English corpora}
\label{tab:HA_ENG_TIME}
\end{table*}

\begin{table*}[!h]
\small
\centering
\begin{tabular}{@{}lrrrrrr@{}}
\toprule
                    & \multicolumn{3}{c}{\textbf{~~~~~~~~~~~~~~~~~~Phase I: Random sample}}                 & \multicolumn{3}{c}{\textbf{~~~~~~~~~~~~~~~~~~Phase II: Model prediction}}        \\
\textbf{Usages}     & \textbf{All} & \textbf{Modern} & \textbf{Historical} & \textbf{All} & \textbf{Modern} & \textbf{Historical} \\ \midrule
\textbf{total}     & 706 & 337 & 369 & 1001 & 478 & 523 \\
\textbf{excluded}  & 52  & 4   & 28  & 74   & 9   & 65  \\
\textbf{remaining} & 674 & 333 & 341 & 927  & 469 & 458 \\
\textbf{assigned}  & 562 & 293 & 269 & 327  & 224 & 103 \\
\textbf{unassigned} & 112 (16.6\%) & 40 (12.0\%)     & 72 (21.1\%)         & 600 (64.7\%) & 245 (52.2\%)    & 355 (77.5\%)       \\ \bottomrule
\end{tabular}
\caption{Annotation results on Swedish corpora}
\label{tab:HA_SWE_TIME}
\end{table*}

\twocolumn

\end{document}